\documentclass[sigconf]{acmart}
\usepackage{algorithm2e}

\setcopyright{acmlicensed}
\copyrightyear{2019} 
\acmYear{2019} 
\acmConference[CIKM '19]{The 28th ACM International Conference on Information and Knowledge Management}{November 3--7, 2019}{Beijing, China}
\acmBooktitle{The 28th ACM International Conference on Information and Knowledge Management (CIKM '19), November 3--7, 2019, Beijing, China}
\acmPrice{15.00}
\acmDOI{10.1145/nnnnnnn.nnnnnnn}
\acmISBN{978-1-nnnnnnn.nnnnnnn}

\begin{document}
\fancyhead{}



\title{ConCET: Entity-Aware Topic Classification for Open-Domain Conversational Agents}

\author{Ali Ahmadvand, Harshita Sahijwani, Jason Ingyu Choi, and Eugene Agichtein}
\affiliation{\institution{Computer Science Department, Emory University, Atlanta, GA, USA}}
\email{{aahmadv, hsahijw, ichoi5, eugene.agichtein} @emory.edu}






\begin{abstract}

Identifying the topic (domain) of each user's utterance in open-domain conversational systems is a crucial step for all subsequent language understanding and response tasks. In particular, for complex domains, an utterance is often routed to a single component responsible for that domain. Thus, correctly mapping a user utterance to the right domain is critical. To address this problem, we introduce ConCET: a Concurrent Entity-aware conversational Topic classifier, which incorporates entity-type information together with the utterance content features. Specifically, ConCET utilizes entity information to enrich the utterance representation, combining character, word, and entity-type embeddings into a single representation. However, for rich domains with millions of available entities, unrealistic amounts of labeled training data would be required. To complement our model, we propose a simple and effective method for generating synthetic training data, to augment the typically limited amounts of labeled training data, using commonly available knowledge bases as to generate additional labeled utterances. We extensively evaluate ConCET and our proposed training method first on an openly available human-human conversational dataset called Self-Dialogue, to calibrate our approach against previous state-of-the-art methods; second, we evaluate ConCET on a large dataset of human-machine conversations with real users, collected as part of the Amazon Alexa Prize. Our results show that ConCET significantly improves topic classification performance on both datasets, including 8-10\% improvements over state-of-the-art deep learning methods. We complement our quantitative results with detailed analysis of system performance, which could be used for further improvements of conversational agents.
\end{abstract}

\maketitle

\section{Introduction}
\label{sec:Introduction}

Open-domain conversational agents increasingly require accurate identification of the topic (domain) and intent of the utterance. Often, domain classification is one of the first steps, and an error in this step can cause a cascading effect throughout the system, and degrade the overall performance. Most current conversational systems use a component architecture \citep{Ram:2017}, where each user utterance is assigned to a domain-specific component such as { \em Movie} Bot{\em \footnote{\em{\url{https://www.amazon.com/Amazon-MovieBot/dp/B01MRKGF5W}}}}. Mis-classifying the intent of an utterance and assigning it to the wrong component, can produce erroneous responses, and degrade the user experience. 





An important challenge for conversational domain classification is that keyword-based classification is not sufficient. Domain-specific keywords or triggers might help for queries like ``Let's talk about my dog'', since the word ``dog'' appears frequently in utterances from the {\em Pets\_Animals} domain. However, they do not enable us to correctly classify utterances containing ambiguous keywords that can refer to multiple entities. 
For example, to correctly classify utterances like ``When is the next Hawks game?'', we need to take into account all the possible types of entities that the word ``Hawks'' might be referring to, i.e. the bird hawk and the sports team Atlanta Hawks, as well as the context, which mentions ``game''. 

Moreover, the creation of new entities, like recent movies, makes the model obsolete with time. To fix this problem, it would be necessary to constantly keep updating the model by incorporating new information about people, organizations, movies and other entities, which can cause unintended effects in the model, and would be inefficient. 
To address these problems, we introduce a novel, data-driven approach to entity-aware conversational topic classification: a deep learning algorithm named Concurrent Entity-aware Topic classifier (ConCET) augmented with external knowledge about entities and their types, retrieved dynamically from a knowledge base, using either a publicly available entity linker, or one fine-tuned for the expected utterances. ConCET combines the implicit and explicit representations of the utterance text, together with the semantic information retrieved about the mentioned entities. 
To train ConCET, we introduce a synthetic dataset, created from the expected entities and entity-types, to augment the limited labeled conversational data. This dataset is modeled to approximate the real human-machine conversations observed with real users, as described below. As a part of this paper's contribution, we plan to release this synthetic dataset to the research community. We evaluate ConCET on an openly available human-human conversational dataset, and a large dataset of human-machine conversations with real users, collected as part of the Amazon Alexa Prize 2018. Our results show that ConCET significantly improves topic classification performance on both datasets, reaching 8-10\% improvements compared to state-of-the-art deep learning methods.

In summary, our contributions are: (1) The development of ConCET, a novel entity-aware topic classifier by combining implicit and explicit representations of an utterance and fusing them with handcrafted features; (2) Incorporating external knowledge about entities retrieved from a knowledge base; and (3)  creation of a new large-scale synthetic yet realistic dataset for training topic classification systems designed for open-domain conversational agents. Next, we present related work to place our contributions in context. 

\begin{figure*}
\centering
\includegraphics[ width=150mm]{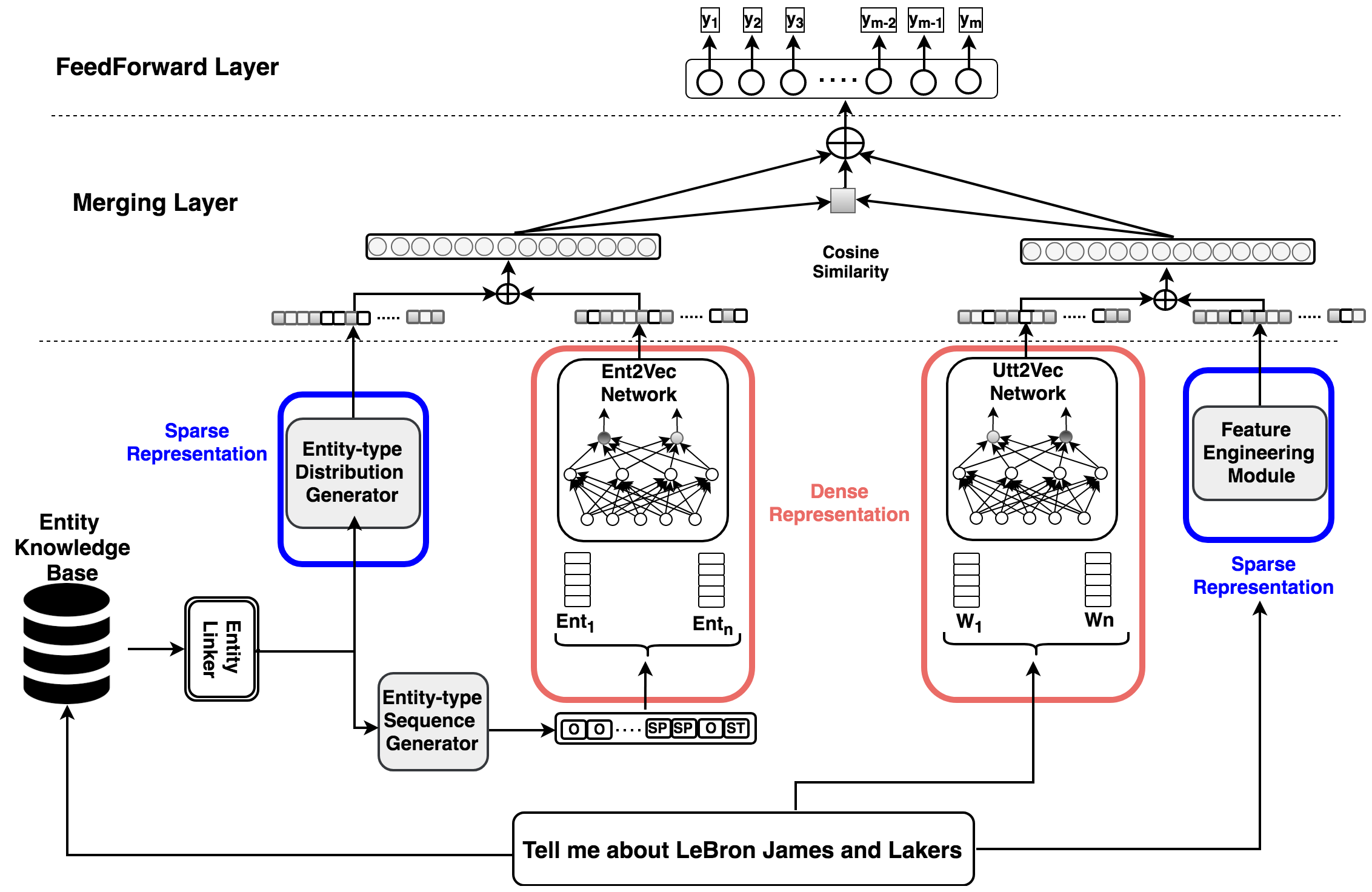}
\caption{The overall network architecture for Entity-Aware Topic Classifier (ConCET) model, where ``SP'' and ``ST'' stand for {\em Sports\_Player} and {\em Sports\_Team} entity-types.}
\label{fig:OveralCET}
\vspace{-3mm}
\end{figure*}

\vspace{-3mm}
\section{Related Work}

In this section, we first describe different types of conversational agents, followed by general text classification methods. We then discuss some existing entity linking approaches, and methods for entity-based text representation. Finally, we discuss the related work in the area of topic classification for conversational agents.  
\vspace{-2mm}
\subsubsection*{\bf Conversational systems. } 
Conversational agents such as Amazon's Alexa, Apple's Siri, and Microsoft's Cortana are becoming increasingly popular. Most existing conversational agents are designed for a single domain like {\em Movie} or {\em Music}. Having an open-domain conversational agent that coherently and engagingly converses with humans on a variety of topics is a long-term goal for dialogue systems ~\citep{guo2018topic, venkatesh2018evaluating, Khatri:2018}. A domain classifier to understand the conversational topics is crucial for the success of an open-domain conversational system.



\vspace{-2mm}
\subsubsection*{\bf Text classification methods. }Topic classification in open-domain dialogue systems can be treated as a text classification problem, although utterance classification is a much more challenging task compared to general text classification due to four main factors: 1) Human utterances are often short; 2) Errors in Automatic Speech Recognition (ASR); 3) Users frequently mention out-of-vocabulary words and entities; 4) Lack of available labeled open-domain human-machine conversation data. Text classification models have traditionally used handcrafted features like bag-of-words, tf*idf, part-of-speech tagging, and tree kernels \citep{Wang:tutorial, Matt:2013}. However, current models put more focus on the semantic and implicit information in the text using features like the word to vector representation described in  ~\citep{Mikolov:2013} and universal sentence encoder proposed in ~\citep{cer2018Daniel}. Wang et al. in ~\citep{Wang:tutorial} classified text classification algorithms into two main categories: implicit representation based models and explicit representation based models. 
Both types of representations have their advantages. The method proposed in this paper, therefore, uses both handcrafted and semantic features of the utterance.

\vspace{-2mm}
\subsubsection*{\bf Entity linking. } 
Entity linkers identify entity mentions in a text, and resolve and map the mentions to a unique identifier in an associated knowledge base; a common choice for which is Wikipedia. 
Babelfy \citep{moro2014entity} uses a graph-based approach for jointly performing word sense disambiguation and entity linking. 
DBpedia Spotlight \citep{mendes2011dbpedia} links entity mentions to their DBpedia \cite{auer2007dbpedia} URIs.
The SMAPH \citep{cornolti2016piggyback} system for linking web-search queries piggybacks on a web search engine to put the query into a larger context, and further uses a supervised ranking model to predict the joint annotation of the full query.

\vspace{-2mm}
\subsubsection*{\bf Entity-based text representation. } Entity-based text representation has been studied in different fields such as information retrieval \citep{BOE2017}, question answering \citep{Distrep2017}, and coherence modeling \citep{entityGrid2018, coherence2018}. Yamada et al. \citep{Distrep2017} proposed a model to encode entity information in a corpus such as Wikipedia into a continuous vector space. This model jointly learns word and entity representations from Wikipedia and DBpedia. \citep{Entshorttext2018} proposed a CNN-based model for merging the text and entities extracted from a large taxonomy knowledge base for short-text classification. 
Moreover, despite these efforts, we are not aware of a published result evaluating entity linking for conversational topic classification, beyond using the utterance content itself~\citep{Khatri:2018}. 
In this paper, we propose a neural network architecture and processing pipeline, for conversational topic classification where the entities, their significance, and positional order are taken into account.

\vspace{-2mm}
\subsubsection*{\bf Deep learning approaches. } 
Both CNN ~\citep{kim:2014,conneau2016very} and RNN ~\citep{rnn2016Pengfei} models show promising results for text classification.
Lecun et al. proposed VDCNN ~\citep{conneau2016very} for text classification based on a popular model in computer vision, which they redesigned for text classification. 
FastText ~\citep{joulin2016fasttext} is another character-embedding based text classification model that was released by Facebook for efficient learning of word representations and text classification. 
Character-embedding based models have shown higher robustness in representing misspellings and out-of-domain words. 
Our approach builds on the success of deep learning models for classification and we explore both CNN and RNN models in our implementation. Moreover, we employ character-based modeling to further make the classification robust to ASR errors and out-of-domain words which are frequent in conversations.
\vspace{-2mm}
\subsubsection*{\bf Conversational topic classification. }
Yeh et al. in ~\citep{Yeh:2014} propose using Latent Dirichlet Allocation (LDA) for topic modeling in dialogues. 
Although LDA-based models are effective for topic modeling, they cannot adapt to users' existing knowledge ~\citep{Mimno:2011}. Guo et al. \citep{guo2018topic} propose Attentional Deep Averaged Network (ADAN), an unsupervised neural network method which learns relevant keywords' saliency for the corresponding topic classes. 

In this paper, we extract both textual and entity information from an utterance and combine them through a deep learning model for conversational topic classification. The new proposed model, called {\bf ConCET}, is an entity-aware topic classifier for open-domain conversational agents. In contrast to ADAN, which uses lexical keyword features, our proposed method uses a dynamic knowledge base to recognize the entities in user utterances and embed that information as part of the representation. We believe that using a dynamic knowledge-based model is more effective than an unsupervised keyword-based method because a model that assumes a static knowledge-base would need to be retrained constantly for new entities, introducing unnecessary model complexity and dependencies.

\vspace{-3mm}
\section{ConCET System Overview} 
\label{sec:CET}


We now introduce our ConCET system at a high level, before diving into implementation details. Our proposed ConCET model is illustrated in Figure \ref{fig:OveralCET}.

ConCET utilizes both textual and entity information from an utterance. To represent textual and entity information, ConCET extracts both sparse and dense representations. To this end, a pipeline of deep neural networks and handcrafted feature extraction modules is designed. This pipeline consists of four components namely Utterance-to-Vector (Utt2Vec) network, feature engineering module, Entity-to-Vector (Ent2Vec) network, and the Entity-type distribution generator. The Entity-type distribution generator module uses an entity linker to get the entity-type distribution corresponding to each entity in the utterance. 

Utt2Vec and the feature engineering module extract the textual representation. Utt2Vec is a deep neural network model which utilizes character, word, and POS tags for utterance representation. Feature engineering module extracts handcrafted features such as LDA and LSA topical distribution from an utterance. Finally, they are combined through a fully-connected neural network.

To model the entity information, ConCET utilizes both the entity-type distribution and the order of entity-types appearing in the utterances.
Ent2Vec network is responsible for mapping this entity sequence representation to a high dimensional vector. Entity-type distribution features and the output of the Ent2Vec network are combined through a fully-connected neural network.

Next, the cosine similarity\footnote{Dot product also can be used. In this case, the entity vector should be normalized to unit length.} between textual and entity representations is computed. This similarity value, concatenated with the textual and entity representations, is fed to a feed-forward layer to compute the final softmax distribution of topics. 

To summarize, ConCET proposes an entity-aware text representation model that learns a ternary representation of character, word and entity information. In the next section, we introduce the entity linking methods used to derive entity-based information. We conducted our experiments using two different entity linkers to measure the sensitivity of the ConCET model to the entity linking step. Then, in Section \ref{sec:CETmodel} we explain the details of the ConCET model.


\vspace{-2mm}
\section{Conversational Entity Linking}
\label{sec:entitylinking}
In this section, we describe the two entity linkers that were used for detecting entities and their type distributions. The type information is used for semantic representation in the ConCET model. 

We emphasize that the focus of this work is {\em not} on developing a novel entity linker, which is an important area of research on its own. Rather, we experiment with an off-the-shelf entity linker, DBPedia Spotlight\footnote{https://github.com/dbpedia-spotlight/spotlight-docker}, and our own PMI-based domain-specific entity linker (PMI-EL), designed to cover in more depth some of the conversation domains and entity-types most relevant to our conversational agent. Our experiments with different off-the-shelf entity linkers during the development of our conversational agent showed these two linkers are the most effective for topics that our bot supported. We describe both entity linkers in depth in the next section, here we want to emphasize that the proposed classifier model can incorporate the output of any available entity tagger or linker. 



\vspace{-2mm}
\subsection{DBpedia Spotlight}
\label{sec:entity}
DBpedia Spotlight annotates DBpedia resources mentioned in the text as described in reference \citep{mendes2011dbpedia}. It annotates DBpedia resources of any of the 272 classes (more than 30 top-level ones) in the DBpedia Ontology. It performs entity annotations in 3 steps, 1) spotting, 2) candidate selection, and 3) disambiguation. It uses the Aho-Corasick string matching algorithm for finding all the phrases which could potentially be entity mentions or surface forms.
It then finds candidate entities for each surface form using the DBpedia Lexical Dataset. 
For disambiguation, each candidate DBpedia resource is first modeled in a Vector Space Model (VSM) as the aggregation of all paragraphs mentioning that concept in Wikipedia. The candidates are then ranked by their $tf*icf$ cosine similarity score with respect to the context, where the $icf$ score estimates how discriminating a word is, which is assumed to be inversely proportional to the number of DBpedia resources it is associated with. 

\vspace{-2mm}
\subsection{PMI-based Entity Linker (PMI-EL)}
We created a domain-specific entity linker called PMI-EL for our conversational system for the Alexa Prize, which annotates the 20 entity-types most relevant to our system. 
It links entities to an associated knowledge base containing all the entities supported by our conversational agent. PMI-EL follows similar steps to DBpedia Spotlight. However, it does not use the utterance context in the disambiguation step and relies solely on an estimated prior distribution of types for a given entity for disambiguation. The main reason was that most of the user utterances were short (average utterance length of 3.07 words), and sometimes consisted of just the entity name. Thus, the context was often not helpful or present, and type inference based on prior probabilities may be sufficient for this setting.
We next describe the process by which the knowledge base was constructed, and how the prior type probabilities were estimated for entity-type inference.

\vspace{-2mm}
\begin{table}
\footnotesize
      \centering
          \begin{tabular}{llll}
           \toprule
         Movie\_Name &Celebrities& Authors &Bands \\ 
         Sports\_Team&Sportname&Companies&Food  \\ 
         Organization&Politicians&Universities&Singers \\ 
         Songname&Animal&Country&Actors\\
         Hotels\_Foodchains&Tourist\_points&Genre\_Books &City\\
          \bottomrule
         \end{tabular}
      \caption{Entity-types recognized by PMI-EL.}
      \label{tab:entitytypes}
      \vspace{-0.5cm}
\end{table} 

\vspace{-2mm}
\subsubsection*{\bf PMI-EL knowledge base construction}
Our knowledge base starts with entities from a snapshot of DBpedia from 2016. Additionally, to provide coverage of current entities of potential interest to the user, we augment the knowledge base by adding entities that our open-domain conversational agent supports. We periodically retrieve entities from the following sources and domains:
\begin{itemize}
    \item Persons, Organizations and Locations: from news provided by Washington Post{\em \footnote{\em{\url{https://www.washingtonpost.com/}}}}
    \item Cities and Tourist Attractions: from Google Places API{\em \footnote{\em{\url{https://developers.google.com/places/web-service/search}}}} 
    \item Bands and Artists: from Spotify{\em \footnote{\em{\url{ https://www.spotify.com/us/}}}} and Billboard{\em \footnote{\em{\url{https://www.billboard.com/}}}} 
    \item Books and Authors: from Goodreads{\em \footnote{\em{\url{ https://www.goodreads.com/}}}} and Google Books{\em \footnote{\em{\url{ https://developers.google.com/books/}}}}
    \item Actors and Movies: from IMDb{\em \footnote{\em{\url{https://www.imdb.com/}}}} and Rotten Tomatoes{\em \footnote{\em{\url{https://www.rottentomatoes.com/}}}}
\end{itemize}

We maintain an index of all the entities and their corresponding types using ElasticSearch{\em \footnote{\em{\url{ https://www.elastic.co/products/elasticsearch}}}}, which is used in the online entity linking step.

\vspace{-2mm}
\subsubsection*{\bf PMI-based type distribution}
For entities with more than one type, we also index the estimated pointwise mutual information (PMI) \citep{bouma2009normalized}
of the entity with all its types. 
PMI is a measure of how much the actual
probability of a particular co-occurrence of events p(x, y) differs from what we would expect it to be on the basis of the probabilities of the individual events and the assumption of independence of events $x$ and $y$, and is calculated as:
\begin{equation}
PMI(x, y) = ln\left(\frac{p(x, y)}{p(x)p(y)}\right)
\end{equation}

To predict the most likely type for entities with multiple types, we estimate the point-wise mutual information (PMI) of the entity with each type by counting the co-occurrences of the entity and the type's name in a large corpus, which has been shown to correlate with the probability of association \citep{bouma2009normalized}. 
More formally, the entity-type PMI score is computed as:\\
\begin{gather}
    PMI(m, t_i) = \frac{|(Docs(m,C) \cap Docs(t_i,C))|}{|Docs(m,C)|}
\end{gather}

where $m$ is an entity mention, $t_i$ is a type, $C$ is a corpus and $Docs(phrase, C)$ is a set of documents in $C$ containing a given phrase. 
For our experiments, as $C$ we used a publicly available corpus of 46 million social media posts from a snapshot of Reddit.\footnote{\em \url{https://files.pushshift.io/reddit/submissions/}}

For example, to disambiguate the mention ``Kings'' in a user's utterance, we compute the number of times each type name co-occurred with the word ``Kings'' in the corpus, normalized by the total number of occurrences of the word ``Kings'' itself. 
In this example, the type distribution for a string ``Kings'' is: $[Sports\_Team: 0.54, Movie\_Name: 0.44, City: 0.02]$. 
Because of the large size and diversity of the corpus, PMI is expected to be a good estimate of type distribution. Despite potential noise in estimating type distribution for some polysemous entities,
the ConCET model is able to use the type distribution, as we demonstrate empirically under a variety of conditions.

\subsubsection*{\bf PMI-EL entity detection in utterances}
To support efficient entity linking at run-time, an inverted n-gram entity index was constructed for all entities in the knowledge base. At runtime, entities are detected via n-gram matching against an entity index. For example, if the utterance is ``who won the Hawks and Kings game'', we query the index for ``the Hawks'', ``Kings'', ``Hawks'' and every other possible n-gram with less than 6 words. For this utterance, the response from the entity index would be the entities and the type distributions associated with them, e.g. ``Hawks'': $[Sports\_Team: 0.88, Animal: 0.11, City: 0.01]$ and ``Kings'': $[ Sports\_Team: 0.54, Movie\_Name: 0.44, City: 0.02]$.

The entity detection step has time complexity $O(n^2)$ in the number of words in the utterance since we perform $O(1)$ look-ups for $O(n^2)$ n-grams for each utterance. 
The running time for entity linking is 16 ms on an average for utterances with 4 words which were common, and 100 ms for utterances with 32 words, which were among the longest utterances we encountered. However, PMI-EL would not be efficient if used on very long text.

The output from the entity linker is passed to the Entity Representation Model, described in \ref{sec:entityrep}, which converts it into a suitable representation for the ConCET model.






\section{ConCET: Concurrent Entity-Aware Topic Classifier}
\label{sec:CETmodel}
In this section, we present the details of ConCET model. First, Section \ref{sec:textrep}, describes our model for the textual representation of the utterance. Then, Section \ref{sec:entityrep}, presents the proposed entity representation model. Finally, Section \ref{merging-layer} discusses the merging and decision layer of the ConCET model.

\vspace{-2mm}
\subsection{Textual Representation}
We use character, word, and POS tagging to model the textual representation. Then, we enrich the representation with the unsupervised topic distribution, as described in detail next.

\label{sec:textrep}
\subsubsection*{\bf Utterance to vector (Utt2Vec) network}
\label{utt-vec}
Utt2Vec network takes word tokens \textit{Utt\textsubscript{w}}, characters \textit{Utt\textsubscript{c}} and POS tags \textit{Utt\textsubscript{p}} of an utterance \textit{Utt} as inputs:

\begin{gather}
Utt_w = [w_1; w_2; w_3 \ ... \ w_n] 
\\
Utt_c = [[c\textsubscript{11} ... c\textsubscript{1k}]; [c\textsubscript{21} ... c\textsubscript{2k}]; \ ... \ [c\textsubscript{n1} ... c\textsubscript{nk}]] 
\\
Utt_p = [p_1; p_2; p_3 \ ... \ p_n] 
\end{gather}

\begin{figure}[ht]
\includegraphics[width=80mm]{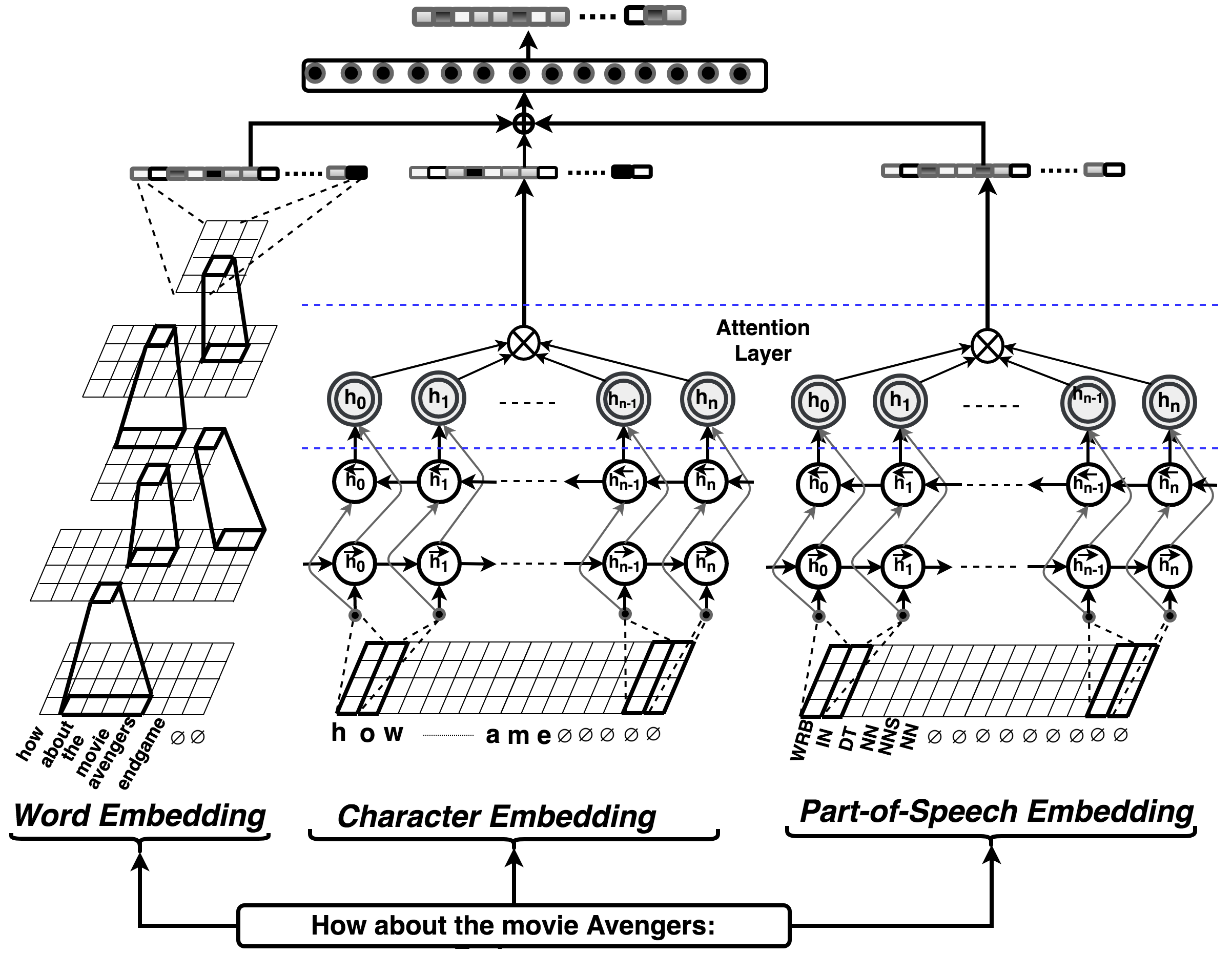}
\caption{Utt2Vec network.}
\label{fig:CNN}
\end{figure}

We use the NLTK\footnote{\em{\url{http://www.nltk.org}}} library for extracting POS tags. Utt2Vec network allows freedom of combining different deep learning architectures such as CNN and RNN to extract features. We define three functions \textit{f\textsubscript{w}, f\textsubscript{c},} and \textit{f\textsubscript{p}} that each take these inputs and output learned hidden representations (h):\\
\begin{gather}
 h_w = f_w(Utt_w) \\
 h_c = f_c(Utt_c)\\
 h_p = f_p(Utt_p) 
\end{gather}

For our implementation, $f\textsubscript{w}$ is a 3-layered CNN with max pooling. For $f\textsubscript{c}$ and $f\textsubscript{p}$, we use 1-layered BiLSTM network with global attention. For the word embedding layer, we pre-initialize the weights using Word2Vec vectors with size 300. The weights on the word embedding layer are tuned during training. For character and POS embeddings, we randomly initialize the embedding layer with size 16. Given the hidden representations of each timestamp $h\textsubscript{i}$ in LSTM cells, dot product similarity score s\textsubscript{i} is computed based on a shared trainable matrix \textit{M}, context vector \textit{c} and a bias term \textit{b\textsubscript{i}}. Softmax activation is applied on similarity scores to obtain attention weights \( \alpha \). Lastly, using learned \( \alpha \), weighted sum on BiLSTM hidden representations is applied to obtain the output \( \hat{h} \) as follows:
\begin{gather}
s_i = tanh(M\textsuperscript{T}h_i + b_i) 
\\
\alpha_i = \frac{exp(s\textsubscript{i}\textsuperscript{T}c)}{\sum_{i=1}^{n}exp(s\textsubscript{i}\textsuperscript{T}c)} 
\\
\hat{h} = \sum_{i=1}^{n} \alpha\textsubscript{i} h\textsubscript{i} 
\end{gather}

\textit{M}, \textit{c}, and \textit{b} are randomly initialized and jointly learned during training. The three outputs from word-CNN $(h\textsubscript{w})$, char-BiLSTM \( \hat{(h\textsubscript{c}}) \), and POS-BiLSTM \( \hat{(h\textsubscript{c}}) \) are concatenated to produce Utt2Vec output:
\begin{gather}
     Utt2Vec\textsubscript{out} = [h\textsubscript{w}; \hat{h\textsubscript{c}}; \hat{h\textsubscript{p}}] 
\end{gather}

This final output is fed to a linear layer of size 256 with ReLU activation and a dropout rate of 0.5 to obtain the utterance vector.







\subsubsection*{\bf Feature engineering module}
The goal of this module is to provide the flexibility of incorporating various external features in ConCET. Since we are focusing on domain classification, we extract unsupervised topic modeling features. However, depending on the data and the task, any type of feature extraction pipeline can be incorporated here. We combined two different topic modeling algorithms, LDA and LSA, and implemented models using the Gensim library\footnote{\em \url{https://radimrehurek.com/gensim/}}. Given hyperparameter \textit{n}, these models output the unsupervised topic distribution of size $n$. By concatenating the two outputs described in the table below, we obtain a topic distribution vector of size $2n$.

\begin{center}
\footnotesize
    \begin{tabular}{l|l}
    \toprule
    \bf  Features & \bf Short Description \\
    \bottomrule
        \textit{F}\textsubscript{LDA} & LDA topic distribution \\
        \textit{F}\textsubscript{LSA} & LSA topic distribution \\
    \bottomrule
    \end{tabular}
    \label{lda-lsa}
\end{center}

The outputs of these two vectors are concatenated to produce \textit{F\textsubscript{out}}:
\begin{gather}
     F\textsubscript{out} = [F\textsubscript{LDA}; F\textsubscript{LSA}] 
\end{gather}




\subsection{Entity Representation}
\label{sec:entityrep}
We now describe how we encode the entity information from the linker as input to our model. We have two modules to do this encoding, 1) Entity-type sequence generator and 2) Entity-type distribution generator. Entity-type sequence generator converts the input word sequence to an entity-type sequence so that the model can learn to predict the topic based on the order in which different entity-types appeared in the utterance. This sequence is fed into the Ent2Vec network, which creates a high-dimensional vector representation for the sequence. The Entity-type distribution generator constructs an overall entity-type distribution for the utterance by aggregating type distributions for all the entities. Finally, the output of Ent2Vec is concatenated with the entity-type distribution to generate the final entity representation.
We now describe these modules in detail.



\vspace{-2mm}
\subsubsection*{\bf Entity-type sequence generator}
\label{sec:entseqgen}

The input of this module is the list of entities and their type distributions derived from the entity linker. To generate this entity sequence, we need to assign the best type corresponding to each entity. The words that are not a part of an entity are assigned { \em Other} or {\em O}. For example, for ``who won the Hawks and Kings game'', a possible entity sequence vector would be [``who''/{\em O}, ``won''/{\em O}, ``the''/{\em ST}, ``Hawks''/{\em ST}, ``and''/{\em O}, ``Kings''/{\em ST}, ``game''/{\em O }]. However, different entity linkers can differently assign entity-types to each word. Consequently, the resulting entity vector has the exact length of the utterance. 
\begin{gather}
\label{eq:ent-type-rep}
 Utt\textsubscript{ent} = [e_1; e_2; e_3 \ ... \ e_n] 
\end{gather}


\vspace{-2mm}
\subsubsection*{\bf Entity-type distribution generator}
\label{sec:entdistgen}

For this module, we first have to determine the total number of entity-types that we want the model to support. For example, for the PMI-based linker, we support 20 types, and for DBpedia Spotlight, we support the 1000 most frequent entity-types from the training set. After determining the size, the distribution value for each entity-type is either 0, or the maximum value for that type in the list of entity-type distributions.
For the example from the previous section, ``who won the Hawks and Kings game'', the type distributions for the two entities from the PMI-based linker are, respectively, $[Sports\_Team: 0.88, Animal: 0.11, City: 0.01]$ for ``Hawks'', and $[ Sports\_Team: 0.54, Movie\_Name: 0.44, City: 0.02]$ for ``Kings''. In that case, if the entity linker identifies 20 types in total, the final entity-type distribution is $[Sports\_team: 0.88, Movie\_Name: 0.44, Animal: 0.11, City: 0.02]$. The value corresponding to the remaining types in Table \ref{tab:entitytypes} is 0.0 in the final output vector of length 20. 

\vspace{-1mm}
\subsubsection*{ \bf Entity to vector (Ent2Vec) network}
\label{sec:ent2vec}
The input to Ent2Vec network is a list of resolved entity-types per word for \textit{Utt\textsubscript{ent}} from entity-type sequence generator:
\vspace{-0.2cm}
\begin{gather}
     Utt_w = [w_1; w_2; w_3 \ ... \ w_n] \\
 Utt\textsubscript{ent} = [e_1; e_2; e_3 \ ... \ e_n] 
\end{gather}

We define a function \textit{f\textsubscript{e}} that takes \textit{Utt\textsubscript{ent}} and outputs learned hidden representations as follows:
\begin{gather}
     h_e = f_e(Utt\textsubscript{ent}) 
\end{gather}

We also use 1-layered BiLSTM network as our \textit{f\textsubscript{e}} function. We randomly initialize an entity embedding layer that has 16 trainable weights per each entity-type. Then, the same attention mechanism as in Section \ref{utt-vec} is applied to \textit{h\textsubscript{e}} to obtain \( \hat{h}_e \) or Ent2Vec\textsubscript{out}. Lastly, entity-type distribution \textit{Ent\textsubscript{dist}} is concatenated with \( \hat{h}_e \) to obtain the final Entity output:
\begin{gather}
     Ent\textsubscript{out} = [Ent2Vec\textsubscript{out}; \textit{Ent\textsubscript{dist}}] 
\end{gather}
This output is fed to a linear layer of size 100 with ReLU activation and a dropout rate of 0.5 to obtain the final entity vector. 

\vspace{-2mm}
\subsection{Merging and FeedForward Layer}
\label{merging-layer}
We obtained the three different outputs each from Utt2Vec network, feature engineering module and Ent2Vec network. \textit{Utt2Vec\textsubscript{out}} is first concatenated with \textit{F\textsubscript{out}} to obtain the following final textual representation \textit{Text\textsubscript{out}} of an utterance:
\begin{gather}
     \textit{Text\textsubscript{out}} = [\textit{Utt2Vec\textsubscript{out}};\textit{F\textsubscript{out}}] 
\end{gather}

We feed \textit{Text\textsubscript{out}} to a linear layer of size 100 with ReLU activation to obtain vector of the same length as \textit{Ent\textsubscript{out}}.
Cosine similarity between these two vectors are computed and concatenated to obtain 201-dimensional \textit{ConCET\textsubscript{out}}:
\begin{gather}
     \textit{ConCET\textsubscript{out}} = [\textit{Ent\textsubscript{out}}; \textit{Text\textsubscript{out}}; Cos(\textit{Ent\textsubscript{out}}, \textit{Text\textsubscript{out}})] 
\end{gather}

According to \cite{Distrep2017}, cosine similarity represents the normalized likelihood that entity-type  \textit{Ent\textsubscript{out}} appears in \textit{Text\textsubscript{out}}.
Finally, softmax activation is applied to generate a probability distribution over \textit{n} possible domains. 


\vspace{-2mm}
\section{Conversational Dataset Overview}
\label{sec:Dataset}
In this section, we describe the conversational data collected during the 2018 Alexa Prize and another publicly available dataset called Self-Dialogue. We also describe the algorithm we designed to generate synthetic training samples, which will be used to augment the original data.

\vspace{-2mm}
\subsection{Amazon Alexa Prize 2018}
The data for evaluation of the proposed models is collected from the 2018 Alexa Prize, a competition held by Amazon every year since 2017 to advance conversational AI. Our team was one of the 8 semi-finalist teams funded by Amazon for the competition. Users were asked to talk to our conversational bot and give a rating from 1.0 to 5.0 (inclusive) based on their experience. 

\vspace{-2mm}
\subsection{Obtaining True Labels for Alexa Data}

Two hundred conversations from the Alexa Prize data were randomly chosen, which consist of 3,000 utterances and responses. These utterances were manually labeled by three different human annotators, whom we call annotator A, B, and C. The matching and kappa scores between the annotator pairs (A, B), (A, C), and (B, C) are (0.82, 0.78), (0.72, 0.65), and (0.80, 0.75), respectively. Overall, these metrics indicate {\em substantial agreement} between all annotators. The final true labels were selected by majority voting. When there was no majority, one of the labels was randomly selected. The final distribution of annotated topics is shown in Table \ref{tab:alexadist}.

\begin{table}[htbp]
\footnotesize
      \centering
          \begin{tabular}{l l|l l|l l}
           \toprule
          Movie &31\%&Music&20\%& News&16\%  \\ 
          Pets\_Animal&6\%&Sci\_Tech&6\%&Sports&6\% \\ 
          Travel\_Geo&2.5\%&Celebrities&2.5\%&Weather&1.5\% \\ 
          Literature&1.5\%&Food\_Drinks&1.5\% &Other&1.5\% \\
          Joke&1\%&Fashion&1\%&Fitness&1\%  \\
          Games&1\% &&\\
          \bottomrule
         \end{tabular}
      \caption{Topics distribution in Alexa 
      Data.}
      \label{tab:alexadist}
       \vspace{-0.5cm}
\end{table}

We randomly selected 90 conversations for training and 10 conversations for validation. The remaining 100 conversations were reserved for evaluation.



\vspace{-2mm}
\subsection{Self-Dialogue Dataset}
Self-Dialogue dataset\footnote{{\em \url{https://github.com/jfainberg/self_dialogue_corpus}}} released by one of the Alexa Prize teams \citep{edina2018} is a human-human conversational dataset collected by using Amazon Mechanical Turk. 
Given a predefined topic, two workers talked about anything related to this topic for 5 to 10 turns. Although this dataset is not comprised of human-machine conversations, it is one of the few publicly available datasets which has a very similar structure to real human-machine conversations, except that the utterances are syntactically richer. This dataset contains 24,165 conversations from 23 sub-topics and 4 major topics: {\em Movie}, {\em Music}, {\em Sports}, and {\em Fashion}. 
The topic distribution for the Self-Dialogue dataset is 41.6\%, 35.1\%, 22.2\%, and 1.1\% for {\em Movie}, {\em Music}, {\em Sports}, and {\em Fashion}, respectively.



For training, all subtopics are merged into the 4 major topics. We also filtered 198 conversations that were designed only for transitions
from {\em Movie} to {\em Music} topics and 216 conversations with mixed  {\em Movie} and {\em Music} labels because we could not assign a unique label. In addition, some of the utterances in the dataset are non-topical chit-chat utterances. They are mostly used for conversational follow-ups such as {\em Yes-Answers}, {\em Backchannel}, and {\em Conventional-opening}.
Since these utterances are unrelated to domain classification, we removed these types in both the training and the test set. To do this, we annotated all the utterances using pre-trained ADAN \citep{Khatri:2018} classifier, which supports 25 topical domains and one {\em Phatic} domain. The {\em Phatic} domain represents all chit-chat and non-topical utterances and any utterance annotated as {\em Phatic} is removed from both the training set and the test set. To verify the accuracy of ADAN classifier, we randomly selected 20 conversations and asked one human annotator to label each utterance as {\em Phatic} or {\em Non-Phatic}. Based on this setup, inter-annotator agreement of 0.87 and Kappa score of 0.82 were achieved, indicating substantial agreement. The final processed dataset consists of 23,751 conversations (363,003 utterances) on 4 main topics. Finally, we divided the dataset into 70\%, 10\% and 20\% for training, validation, and evaluation, respectively. 

A summary of the Alexa data and Self-Dialogue dataset statistics is reported in Table~\ref{tab:dataset-stats}. Utterances from the Alexa data are significantly shorter (3.07 words on average compared to 9.79 in Self-Dialogue), indicating that often entities may be mentioned without extensive context, e.g., as a response to a system question. 

\vspace{-0.3cm}
\begin{table}[h!]
\footnotesize
\centering
    \begin{tabular}{l|c|c|c}
    \toprule
    \bf  Dataset& Words per & Turns per  & Vocabulary \\
                & Utterance & Conversation & Size\\
    \bottomrule
        Alexa         & 3.07&  16.49 &  16,331               \\
        Self-Dialogue & 9.79  & 5.84 & 117,068   \\
        
    \bottomrule
    \bottomrule
    \end{tabular}
    \caption{Alexa and Self-Dialogue data statistics.}
    \label{tab:dataset-stats}
     \vspace{-1cm}
\end{table}

\subsection{Synthetic Training Data Generation}
We propose a simple yet effective approach to generate many synthetic utterances for training topic classification models. As we will show, this ability can be particularly useful for augmenting real data when limited manual labels are available, to train deep neural network models which require large amounts of labeled training data. The approach is summarized in Algorithm~\ref{synthetic-data-code}. 

For each topic, a small number of predefined intent templates are created. These templates are designed by engineers who developed each domain-specific module. The rules described in Amazon Alexa developers' guide\footnote{\em {https://developer.amazon.com/docs/custom-skills/best-practices-for-sample-utterances-and-custom-slot-type-values.html}} were applied in order to capture the most common topic-specific intents and accommodate enough lexical and syntactic variations in the text. The templates contain slots to be filled with either entities or keywords, for example, ``Play a {\em KEYWORD\_MUSICGENRE} music from {\em NER\_SINGER}'' and ``tell me some {\em KEYWORD\_MOVIEGENRE} films played by {\em NER\_ACTOR}''. 
Each slot starting with NER is filled by an entity from the knowledge base, and each slot starting with KEYWORD is filled using a predefined list of intent-oriented keywords.
For instance, the slot {\em KEYWORD\_MUSICGENRE} is randomly filled using a list of popular music genres like {\em rock}, {\em pop} and {\em rap}.  
We first generated these predefined keywords manually and expanded the lists with the 10 most similar words from WordNet\footnote{\em \url{ https://wordnet.princeton.edu}} for each keyword.
To fill in the entity slots, we used the corresponding lists from our knowledge base (described above), prioritizing the most popular entities, and the most common templates according to domain knowledge and most frequent utterance statistics. While the possible number of generated utterances is the direct product of the number of templates, keyword values, and entity-values, the process ends after a predefined number of synthetic utterances is reached. For our experiments, we control the size of the synthetic dataset with a parameter named $\rho$. This value is determined based on the number of available templates for a topic, importance of a topic, and the overall number of covered topics. We conducted an experiment on this value described in Section \ref{section-detailed-results}. We decided to choose $400K$ to make a trade-off between time and accuracy and to make the experiments manageable.

\begin{table}[htbp]
\footnotesize
\fontsize{6pt}{6pt}
      \centering
          \begin{tabular}{l l|l l|l l}
           \toprule
          Movie &28\%& Music&15\%&Pets\_Animal&13\% \\ 
          Travel\_Geo&12\% &News&10\%&Games&10\%\\ 
          Sports&5\%&Sci\_Tech&3\%&Celebrities&2.5\% \\ 
          Fashion&1\%&Weather&1\%&Literature&1\% \\
          Food\_Drinks&0.9\%&Other& 0.1\%&&\\
          \bottomrule
         \end{tabular}
      \caption{Topics distribution in Synthetic Dataset.}
      \label{tab:syntopics}
      \vspace{-0.9cm}
\end{table} 

Any other external dataset can be incorporated into the synthetic generator above to enrich classes lacking sufficient samples. In our experiments, we did not have as many entities from {\em Technology} and {\em Sports} domain compared to {\em Movies} and {\em Music} domains. Hence, we used an open-source Yahoo-Answers question-answer corpus to add questions for these classes. Since human-machine utterances tend to be short, as reported in Table~\ref{tab:dataset-stats}, we only added questions shorter than 10 words. The final topic distribution of the synthetic dataset is shown in Table. \ref{tab:syntopics}.

\vspace{-2mm}
\begin{algorithm}[ht]
\small
\SetAlgoLined
  
  \textbf{Template and Entity-based Synthetic Utterance Generator}\\ 
  \For{topic in topic\_list}{
     \For{template in common\_topical\_templates}{
      tmp = read(template);\\
      $\rho$ = SYNTHETIC\_DATASET\_SIZE;\\
      \em{e.g. tmp = \scriptsize{``Fun facts for NER\_ANIMALS''}}\\
      \em{e.g. tmp = \scriptsize{``The best KEYWORD\_LEAGUE team''}}\\
      
      slot\_list = find(slots);\\
      \For{entity\_type \textbf{and} keyword\_type in slot\_list }{
          \For{\scriptsize entity in entity\_type \textbf{and} keyword in keyword\_type}{
              
                \eIf{ entity \textbf{in} common\_entity\_list \textbf {and} keyword \textbf{in} common\_keyword\_list}
                {generate\_utterance(temp);\\
                generate\_label();\\
                \eIf{len(dataset) > $\rho$}{return \textbf{ dataset};}{}
                }
                {continue;}
                   
          }
       }
     }
 }
\caption{\textbf{Algorithm to generate the synthetic template-, keyword-, and entity-oriented utterances.}}
\label{synthetic-data-code}
\end{algorithm}
\section{Experimental Setup}
In this section, we first describe baseline methods in Section \ref{sec:Baseline}. Experimental metrics and procedures are described in Section \ref{sec:Exp-metric}. 
All experiments were implemented in Python 2.7 using TensorFlow 1.12.0{\em \footnote{{\em\url{https://www.tensorflow.org}}}} library.

\vspace{-1mm}
\subsection{State-of-the-Art Baselines}
\label{sec:Baseline}
Three state-of-the-art methods were used as baselines:
\begin{itemize}
\item {\bf ADAN} \citep{Khatri:2018}: ADAN was proposed by Amazon for conversational topic classification, and it was trained on over 750K utterances from internal Alexa user data for 26 topics. 
\item {\bf FastText}\citep{joulin2016fasttext}: FastText is a text classification model from Facebook Research. FastText operates on character n-grams and uses a hierarchical softmax for prediction, where word vectors are created from the sum of the substring character n-grams.
\item {\bf VDCNN} \citep{conneau2016very}: This model was proposed as a character-based text classification model. VDCNN, like FastText, can model misspelled words (potentially mitigating ASR problems in human-machine conversations) more robustly than word-embedding based models.
\end{itemize}


\vspace{-0.2mm}
\subsection{Training Parameters}

To train the ConCET model, the parameters for CNN and BiLSTM described in Figure \ref{fig:CNN} were chosen based on our experience and previous literature. Finally, we trained the overall model with an Adams optimizer and a learning rate of 0.001.  
All experiments for ADAN were conducted using the topic classifier API made available to the teams by the Amazon Alexa Prize \citep{Khatri:2018}. To train the FastText model\footnote{\em \url{https://fasttext.cc}}, character 5-grams with word embedding of size 300 were used. Finally, VDCNN results are reported based on a publicly available implementation.\footnote{\em \url{ https://github.com/zonetrooper32/VDCNN}}. The results are reported for a 29-layer VDCNN, based on the original paper.

\noindent{\bf {Evaluation metrics.}}
\label{sec:Exp-metric}
We used two standard classification metrics, {\bf Micro-Averaged Accuracy} and {\bf Micro-Averaged F1} \cite{mitchell2006discipline}, to evaluate our approach. 

\vspace{-0.2cm}
\section{Results and Discussion}
\label{sec:results}
We begin this section by reporting the performance of ConCET in comparison to the baseline models described in Section \ref{sec:Baseline}. Then, we illustrate the impact of the entity, external, and utterance features through a feature ablation study.

\vspace{-0.2cm}
\subsection{Main Results}
\label{sec:main-results}
Table \ref{tab:overallAlexa} summarizes the performance of the models on Alexa and Self-Dialogue datasets. 
The results show that both variations of ConCET outperform the state-of-the-art classifier baselines Fastext, VDCNN, and ADAN on Alexa dataset by large margins of 13\%, 23\%, and 10\%, respectively in terms of Micro-Averaged F1 score. Among the baselines, ADAN has the best results on the Alexa dataset, while VDCNN achieves the best results on the Self-Dialogue dataset. All the improvements are statistically significant using one-tailed Student's t-test with p-value < 0.05.

\begin{table}[ht]
\footnotesize
\centering

\begin{tabular}{@{}l||ll||ll@{}}
\multicolumn{5}{c}{\hspace{45pt} \bf{Dataset}} \\ 

\multicolumn{1}{c||}{\textbf{Method}}&
\multicolumn{2}{c}{Alexa} & 
\multicolumn{2}{c}{Self-Dialogue} \\

\cline{2-5}
\multicolumn{1}{c||}{}  & \multicolumn{1}{c}{Accuracy} & \multicolumn{1}{c||}{F1} & \multicolumn{1}{c} {Accuracy} & \multicolumn{1}{c}{F1} \\
\bottomrule\bottomrule
 FastText \citep{joulin2016fasttext}&   54.54 & 58.34                &              79.21 & 79.32 \\
ADAN \citep{Khatri:2018}&                  62.01  &  66.10              &              46.64 & 59.66 \\
VDCNN \citep{conneau2016very} &            46.48 & 48.56                &              79.98 & 80.61  \\
ConCET \bf{(S)} &    68.75 \tiny (+10.9\%) & 68.73 \tiny(+4.0\%)      &     \bf{84.58 \tiny(+5.7\%)} &\bf{84.71 \tiny(+5.1\%)}  \\
ConCET \bf{(P)} &  \bf{71.46 \tiny(+15.2\%)} & \bf{71.72 \tiny(+8.5\%)}    &      84.59 \tiny(+5.7\%) & 84.66 \tiny(+5.0\%)  \\
 \bottomrule
\end{tabular}
\caption{Topic classification on Alexa and Self-Dialogue datasets, where (S) stands for Spotlight entity linker and (P) stands for the domain-specific PMI-EL entity linker. The relative improvements over ADAN and VDCNN are shown on the Alexa and Self-Dialogue datasets, respectively.}

\label{tab:overallAlexa}
 \vspace{-0.7cm}
\end{table}


Interestingly, the performance of the VDCNN and ADAN methods switches for the human-machine and human-human datasets, as ADAN relies only on keywords, which is not sufficient for complex human-human utterances, while VDCNN exhibits the worst performance for short human-machine utterances. In contrast, ConCET exhibits robust and consistently high performance on both human-human and human-machine conversations.

\vspace{-2mm}
\subsection{Detailed Performance Analysis}
\label{section-detailed-results}
ConCET is a complex model consisting of different steps built based on deep learning models like CNN and RNN. We performed a comprehensive feature ablation analysis to evaluate the effect of each subsection on the overall performance of the system.

\subsubsection*{\bf { Entity linker evaluation}}
While entity linking is not the focus of this paper, since entities and their types play a central role in our approach, entity linking performance could have a significant effect on the overall classifier performance. To quantify the downstream effects of the entity linking accuracy, and to understand whether ConCET can operate with inaccurate entity linkers, we manually annotated entity-types for 350 utterances, which contained entities spotted by at least one entity linker. The distribution over classes is similar to that indicated in Table \ref{tab:alexadist}, with a higher number of utterances from {\em Movies}, {\em Music}, and {\em Travel\_Geo} compared to the other classes. Table \ref{tab:el-eval} presents the accuracy and F1 values of PMI-EL and Spotlight on different classes of utterances. 

\begin{table}[htb]
\footnotesize
\centering
\begin{tabular}{@{}l||ll||ll@{}}
\multicolumn{5}{c}{ \hspace{65pt} \bf {Entity Linker}} \\ 

\multicolumn{1}{c||}{\textbf{Class}}&
\multicolumn{2}{c}{PMI-EL} & 
\multicolumn{2}{c}{Spotlight} \\

\cline{2-5}

\multicolumn{1}{c||}{}  & \multicolumn{1}{c}{Accuracy} & \multicolumn{1}{c||}{F1} & \multicolumn{1}{c} {Accuracy} & \multicolumn{1}{c}{F1} \\
\bottomrule\bottomrule
Movie&                          \textbf{80.00} & 77.19        &             71.83 & 78.46 \\
Travel\_Geo&           \textbf{80.77} & \textbf{87.50}         &             75.47 & 82.47 \\
Music&          65.51 & 59.37         &             64.44 & \textbf{72.5} \\
Sports&          70.56 & 63.16                   &                \textbf{84.00} & \textbf{91.30}     \\
News &                      \textbf{76.47} & \textbf{78.78}         &             66.66 & 70.59 \\
Others&          \textbf{53.68} & 54.84         &             50.69 & \textbf{62.12} \\
\midrule
Overall&          \textbf{68.30} & 68.18         &             63.48 & \textbf{72.67} \\

 \bottomrule
\end{tabular}
\caption{Accuracy and F1 scores of entity detection by PMI-EL and DBPedia Spotlight entity linkers.}
\label{tab:el-eval}
\vspace{-0.5cm}
\end{table}

The two entity linkers exhibit comparable performance, with PMI-EL showing higher Accuracy on the {\em Movies}, {\em Music}, {\em Travel\_Geo}, and {\em News} topics, but DBpedia Spotlight exhibiting higher overall F1 scores. As we will show later in this section, ConCET can perform well with either entity linker. 

\subsubsection* {\bf {Impact of textual representation}}
To evaluate the impact of the textual representation choices, we conducted a feature ablation study. Table \ref{tab:utt} summarizes the results, which indicate that all of the implemented components are significantly contributing to the final performance. Both Utt2Vec and TopicDist representations contribute to the classification performance, but the contributions are greater in Alexa dataset, due to a stronger correlation between the keywords with the user topics.

\begin{table}[htb]
\footnotesize
\centering
\begin{tabular}{@{}l||ll||ll@{}}
\multicolumn{5}{c}{ \hspace{75pt} \bf {Dataset}} \\ 

\multicolumn{1}{c||}{\textbf{\scriptsize Method}}&
\multicolumn{2}{c}{Alexa} & 
\multicolumn{2}{c}{Self-Dialogue} \\

\cline{2-5}

\multicolumn{1}{c||}{}  & \multicolumn{1}{c}{Accuracy} & \multicolumn{1}{c||}{F1} & \multicolumn{1}{c} {Accuracy} & \multicolumn{1}{c}{F1} \\
\bottomrule\bottomrule
 CNN&   47.59 &  42.93         &  79.61 &  79.73 \\
{CNN+$BiLSTM_{pos}$}&       51.60 &  48.14  &   82.82& 82.75\\
                                  &     \scriptsize(+8.4\%) & \scriptsize(+12.1\%)         &  \scriptsize(+4.0\%) &  \scriptsize(+3.8\%) \\
           
 CNN+$BiLSTM_{char}$&      52.40    & 48.65    &    83.12 &   83.01\\
                                 &     \scriptsize(+10.1\%) & \scriptsize(+13.3\%)&  \scriptsize(+4.4\%) & \scriptsize (+4.1\%) \\
 Utt2Vec&                  54.27 & 50.84& 83.33 & 83.35 \\
                                 &     \scriptsize(+14.0\%) & \scriptsize(+18.4\%)&  \scriptsize(+4.6\%) & \scriptsize(+4.5\%) \\
Utt2Vec+TopicDist &        55.88 & 53.09  &  83.45 & 83.75 \\
                                 &     \scriptsize(+17.4\%) & \scriptsize(+23.6\%)&  \scriptsize(+4.8\%) &  \scriptsize(+5.0\%) \\
 \bottomrule
\end{tabular}
\caption{Topic classification Accuracy and F1 for different textual representations Alexa and Self-Dialogue datasets.}
\label{tab:utt}
\end{table}

\subsubsection* {\bf {Impact of entity-type representation}}
Our model utilizes two variants of entity-type representations, namely entity-type distribution (TypeDist) and entity-type sequence modeling (Ent2Vec). We evaluate both entity representation vectors separately on both Alexa and Self-Dialogue datasets. Moreover, we report the result when different combinations of the entity representations are joined with the Utt2Vec network. Table \ref{tab:Ent} reports the contribution of each entity representation to the final performance. 
While both representations contribute greatly to the classifier performance, the effects are greater in the Alexa dataset, due to the strong correlation between the entity-types and the user topics of interest.

\begin{table}[ht]
\footnotesize
\centering
\begin{tabular}{@{}l||ll||ll@{}}
\multicolumn{5}{c}{ \hspace{100pt} \bf {Dataset}} \\ 

\multicolumn{1}{c||}{\textbf{Method}}&
\multicolumn{2}{c}{Alexa} & 
\multicolumn{2}{c}{Self-Dialogue} \\

\cline{2-5}

\multicolumn{1}{c||}{}  & \multicolumn{1}{c}{Accuracy} & \multicolumn{1}{c||}{F1} & \multicolumn{1}{c} {Accuracy} & \multicolumn{1}{c}{F1} \\
\bottomrule\bottomrule
Utt2Vec&                     54.27 & 50.84        &          83.33 & 83.35        \\
Ent2Vec&                     26.93 & 19.93        &          52.45 & 50.32        \\
TypeDist&                     33.73 & 25.54        &          58.95 & 57.00        \\
Ent2Vec+TypeDist&             35.66 & 26.33        &          60.22 & 57.91         \\
Utt2Vec+Ent2Vec &            60.26 & 57.93        &          84.48  & 84.83        \\
&      \scriptsize(+11.3\%)                &\scriptsize(+14.6\%) & \scriptsize(+1.4\%)  & \scriptsize(+1.5\%)   \\
Utt2Vec+TypeDist &            63.46 & 61.03        &          84.43 & 84.71          \\
&                     \scriptsize(+17.0\%) & \scriptsize(+20.0\%)&  \scriptsize(+1.4\%) &  \scriptsize(+1.6\%) \\
Utt2Vec+TypeDist+Ent2Vec &    64.80 & 61.59        &          84.51 & 84.86           \\
&                     \scriptsize(+17.4\%) & \scriptsize(+23.6\%)&  \scriptsize(+1.4\%) &  \scriptsize(+1.8\%) \\
 \bottomrule
\end{tabular}
\caption{Ablation study for different entity representations.}
\label{tab:Ent}
 \vspace{-0.3cm}
\end{table}

\subsubsection* {\bf {Impact of synthetic dataset on ConCET}}
\begin{figure}
\includegraphics[width=75mm]{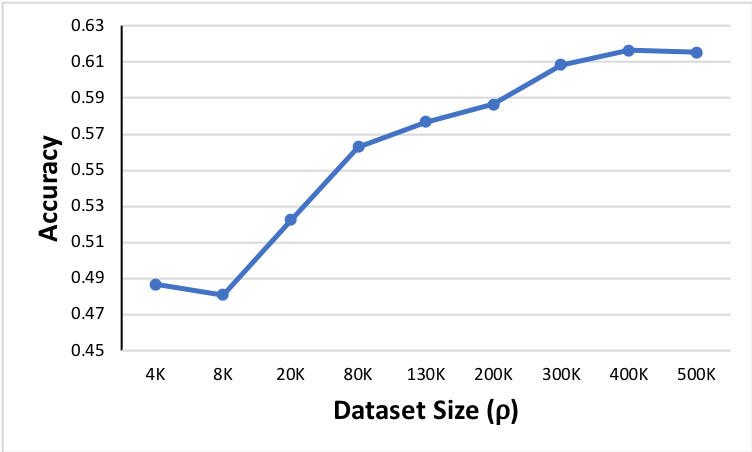}
\vspace{-6mm}
\caption{ConvCET Accuracy on Alexa Prize dataset for varying $\rho$ values in Algorithm \ref{synthetic-data-code}.}
\label{fig:datasetset-size}
\end{figure}

To evaluate the effectiveness of the synthetic dataset, we augmented the Alexa and Self-Dialogue datasets using the synthetic data described above and re-trained the models. The results are reported in Table \ref{tab:synthetic}. Even though the synthetic dataset is effective in the real human-machine conversations with Alexa, it has a negligible impact on the Self-Dialogue dataset. We attribute this effect to the large size of the Self-Dialogue dataset. We argue that even a portion of this dataset is enough for a model to reach its asymptotic performance. To evaluate this hypothesis, we re-trained ConCET in two different settings. First, we randomly sampled 1\% of Self-Dialogue dataset and used it as the training set. Then, we added the synthetic dataset to the sampled portion and trained the model again. In the former case, ConCET reached the Accuracy of $(72.01 \pm 0.1)$, while in the latter case it reached the Accuracy of $(73.12 \pm 0.09)$. We performed each experiment 5 times. This confirms that the size of the labeled dataset is indeed affecting the extent to which the synthetic data can be helpful. 
We conducted an experiment to determine an estimate for the value of $\rho$ using DBPedia Spotlight as the entity linker. The results are shown in Figure \ref{fig:datasetset-size}, which indicate that a value of 400K samples is appropriate for $\rho$ in Algorithm \ref{synthetic-data-code}, due to the classifier peaking at this point with more than 61\% Accuracy.

\begin{table}[ht]
\footnotesize
\centering
\begin{tabular}{@{}l||ll||ll@{}}
\multicolumn{5}{c}{\hspace{100pt} \bf{Dataset}} \\ 

\multicolumn{1}{c||}{\textbf{Train On}}&
\multicolumn{2}{c}{Alexa} & 
\multicolumn{2}{c}{Self-Dialogue} \\

\cline{2-5}

\multicolumn{1}{c||}{}  & \multicolumn{1}{c}{Accuracy} & \multicolumn{1}{c||}{F1} & \multicolumn{1}{c} {Accuracy} & \multicolumn{1}{c}{F1} \\
\bottomrule\bottomrule
Synthetic \bf{(S)}&                          61.60  &  57.44            &            75.62 & 75.52        \\
Synthetic \bf{(P)}&                          62.93  &  63.83            &            58.73 & 59.03                     \\
\\ \hline
Alexa data \bf{(S)} &                        64.81  &  61.92           &               -  & -               \\
Alexa data \bf{(P)} &                        62.93  &  60.24           &               -  & -               \\

Alexa data+Synthetic \bf{(S)} &              68.75 & 68.73     &                - & -               \\
&                     \scriptsize(+6.1\%) & \scriptsize(+10.7\%)&  - &  - \\
Alexa data+Synthetic \bf{(P)} &         \bf {71.46} & \bf {71.72}      &                - & -               \\
&                     \scriptsize(+13.5\%) & \scriptsize(+19.0\%)&  - &  - \\
\\ \hline
Self-Dialogue \bf{(S)} &                            - & -                &           \bf{84.61}  & \bf{85.86}           \\
Self-Dialogue \bf{(P)} &                            - & -                &           84.55  &    84.71           \\
Self-Dialogue+Synthetic \bf{(S)} &               - & -                &      84.58 & 84.71     \\
&                                                - &-                 &  \scriptsize(-0.0\%) &  \scriptsize(-1.3\%)\\
Self-Dialogue+Synthetic \bf{(P)} &               - & -                &             84.59 &  84.66 \\
&                                                - &-                 &  \scriptsize(-0.0\%) &  \scriptsize(-1.4\%)\\
 \bottomrule
\end{tabular}
\caption{Performance of ConCET with and without training on the synthetic dataset, where ``S'' stands for the Spotlight entity linker and ``P'' stands for domain-specific PMI-EL entity linker. }
\label{tab:synthetic}
\vspace{-8mm}
\end{table}

\vspace{-3mm}
\subsection{Discussion}
We now discuss the strengths and potential limitations of the proposed approach. Generally, entity-aware classifiers are prone to overfitting to the majority entity-type. We addressed this difficulty by adding sparse and dense representations of the entity-types, which helps in smoothing the entity representation. In other words, using an additional network and separately training the entities reduced the bias towards entity-types.
Furthermore, there are entity-types like {\em Movie\_Names}, which are notoriously problematic for classification. For example, the utterance ``Fabulous how are you echo'' can be easily mis-classified if the entity-aware model is biased toward certain entity-types. In this example, ``Fabulous'' could be a {\em Movie\_Name}, and ``Echo'' could be a {\em City} located in Oregon. In such cases, the ConCET model avoids this error in two different ways. First, because combinations like these appear in all classes, the classifier tends to be less biased to these entities. Second, two different joint deep network layers are used in ConCET model, which makes the system more robust to entity-type errors.

The ConCET model enriches the textual representation of an utterance with entity information for topic classification. By simultaneously learning the text and entity-types, ConCET captures the likelihood of the appearance of a specific entity-type in an utterance text to thereby learn a specific topic label. Moreover, to model semantic (dense) representations of the entity-types, we computed an entity-type sequence as Equation \ref{eq:ent-type-rep}. The interactions between entity-types, when more than one entity-type appear in the utterance, as well as the order of their appearances in an utterance, can, therefore, be inferred. As a result, ConCET can jointly learn a semantic (dense) representation and the distribution of entity-types with textual information to represent an utterance.



Although ConCET outperforms all of the state-of-the-art baselines with either entity linker, we observe higher improvements on Alexa data with the PMI-EL domain-specific linker. We conjecture that this is because PMI-EL is designed to identify the entity-types supported by the conversational agent, which are better aligned with the target domains. 
Nevertheless, ConCET exhibits significant improvements over the previous state of the art with an off-the-shelf generic entity linker, and, when available, can take advantage of the domain-specific entity linking for additional improvements. 

A reference implementation of ConCET and the associated entity linker implementations, training data, models, and the Knowledge Base snapshot will be released to the research community\footnote{Available at {\em {\url{https://github.com/emory-irlab/ConCET}}}}.

     
   

Deploying a complex system like ConCET in production could potentially degrade system performance by introducing higher response latency. This is an important issue, as response latency has a dramatic effect on the user experience. Interestingly, the classification latency for the proposed approach is not substantially higher compared to the baseline classifier that operates on an utterance text alone. The main reason is that all the 4 stages of the ConCET can be run in parallel. In addition, while entity linking requires a knowledge base lookup, modern in-memory KB storage implementations support candidate entity retrieval and matching in only 10s of milliseconds, which does not introduce perceptible increases to response latency. Finally, ConCET can be executed in parallel for different conversations, allowing the system higher overall throughput without increasing latency for each user. 


\vspace{-2mm}
\section{Conclusions and Future Work}
\vspace{-1mm}
In this paper, we introduced ConCET, a novel and effective entity-aware classifier which fuses textual and semantic entity-oriented information to determine the utterance topic. The results of the extensive experimental evaluation on two different datasets show that ConCET significantly outperforms all the existing state-of-the-art utterance classification models introduced both for generic text and for conversational data.


Our future work includes tuning ConCET model in a more robust way to allow for ASR errors, for example, by relying less on the exact and complete entity detection and instead experimenting with character-based representation models. Another promising direction is to explore other neural network architectures, which may able to incorporate longer contextual dependencies. 
In summary, the presented work advances the state-of-the-art in conversational topic classification and lays the groundwork for future research on open-domain conversational agents.

\vspace{-2mm}
\subsubsection*{\bf {Acknowledgements}} We gratefully acknowledge the financial and computing support from the Amazon Alexa Prize 2018.

\vspace{-2mm}
\bibliographystyle{ACM-Reference-Format}
\bibliography{Reference}
\balance
\end{document}